\documentclass[runningheads]{llncs}\usepackage{graphicx}
\usepackage{algorithm}  
\usepackage{algorithmicx}  
\usepackage{algpseudocode}  
\usepackage{amsmath}  
\usepackage{booktabs}
\usepackage{multirow}
\usepackage{regexpatch}
\usepackage{hyperref}
\usepackage{color}
\usepackage[misc]{ifsym} 

\begin{document}

\title{GT U-Net: A U-Net Like Group Transformer Network for Tooth Root Segmentation}
\titlerunning{GT U-Net: A U-Net Like Group Transformer Network}
\author{Yunxiang Li\inst{1}, Shuai Wang\inst{2}, Jun Wang\inst{3}, Guodong Zeng\inst{4}, Wenjun Liu\inst{1}, Qianni Zhang\inst{5}, Qun Jin\inst{6}, Yaqi Wang\inst{7}$^{(\textrm{\Letter})}$}
\authorrunning{Y. Li et al.}

\institute{Microelectronics CAD Center, Hangzhou Dianzi University, Hangzhou, China \and School of Mechanical, Electrical and Information Engineering, Shandong University, Weihai, China  \and School of Biomedical Engineering, Shanghai Jiao Tong University, Shanghai, China \and sitem Center for Translational Medicine and Biomedical Entrepreneurship, University of Bern, Bern, Switzerland \and School of Electronic Engineering and Computer Science, Queen Mary University of London, London, UK \and Department of Human Informatics and Cognitive Sciences, Faculty of Human Sciences, Waseda University, Tokyo, Japan  \and College of Media Engineering, Communication University of Zhejiang, Hangzhou, China \\ \email{wangyaqi@cuz.edu.cn} }

\maketitle
\begin{abstract}
To achieve an accurate assessment of root canal therapy, a fundamental step is to perform tooth root segmentation on oral X-ray images, in that the position of tooth root boundary is significant anatomy information in root canal therapy evaluation. However, the fuzzy boundary makes the tooth root segmentation very challenging. In this paper, we propose a novel end-to-end U-Net like Group Transformer Network (GT U-Net) for the tooth root segmentation. The proposed network retains the essential structure of U-Net but each of the encoders and decoders is replaced by a group Transformer, which significantly reduces the computational cost of traditional Transformer architectures by using the grouping structure and the bottleneck structure. In addition, the proposed GT U-Net is composed of a hybrid structure of convolution and Transformer, which makes it independent of pre-training weights. For optimization, we also propose a shape-sensitive Fourier Descriptor (FD) loss function to make use of shape prior knowledge. Experimental results show that our proposed network achieves the state-of-the-art performance on our collected tooth root segmentation dataset and the public retina dataset DRIVE. Code has been released at~\href{https://github.com/Kent0n-Li/GT-U-Net}{https://github.com/Kent0n-Li/GT-U-Net}. 
\end{abstract}

\section{Introduction}\label{sec:intro}
In worldwide, approximately 743 million people are affected by severe periodontitis, which is considered the sixth most common health disorder~\cite{peres2019oral}. Nowadays, root canal therapy is a routine periodontal treatment for periodontitis and an incorrect evaluation of the treatment result will impede timely follow-up~\cite{estrela2014characterization,kaplan2019dental}. Since the tooth root boundary is the significant anatomy feature for carrying out the evaluation, tooth root segmentation becomes the most important step of root canal therapy automatic evaluation. Unfortunately, performing an accurate tooth root segmentation is a very challenging task due to the following reasons: 1) the tooth root boundaries are blurry and some tissues around the teeth have similar intensities to the teeth, as shown in Fig. 1 (a); 2) other bones and tissues may overlap with the tooth root in the oral X-ray images, as shown in Fig. 1 (b); 3) the quality of the X-ray image may be very poor such as overexposed or underexposed, as shown in Fig. 1 (c).
\begin{figure}[ht]
  \centering
  \includegraphics[width=1\textwidth]{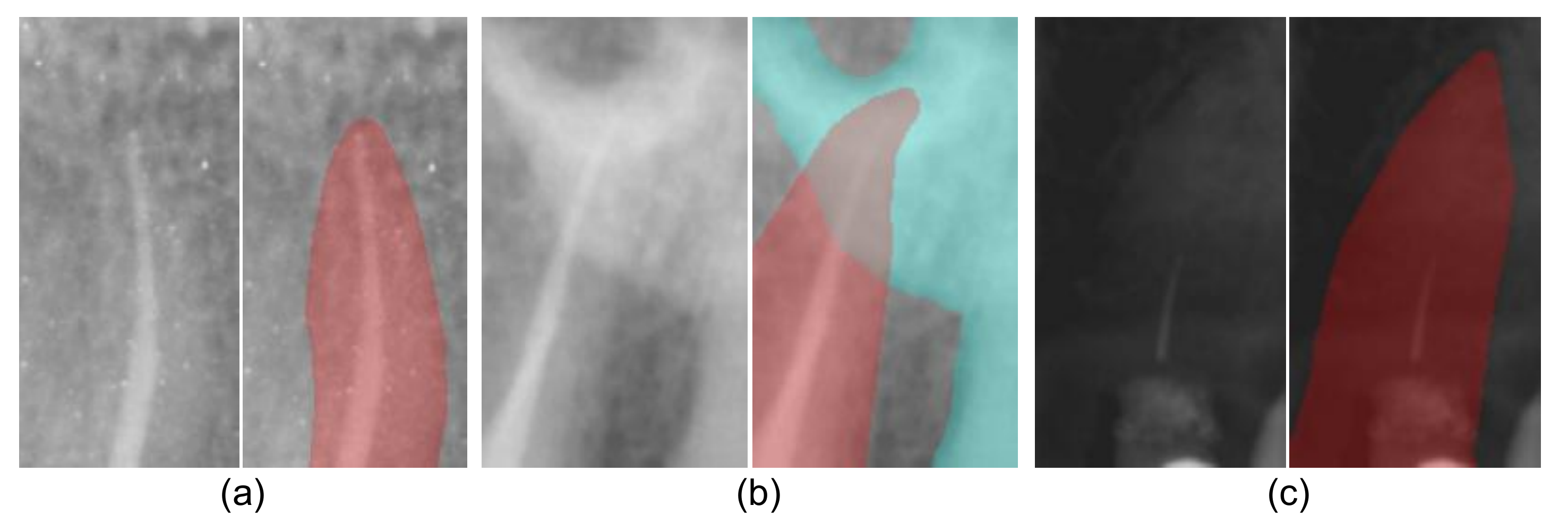}
  \caption{Three examples of root canal therapy X-ray images. The left image in each group is the original image, and the right one depicts the tooth root and other tissues by red and cyan, respectively.}
  \label{fig:fig01}
\end{figure}

To address the problems mentioned above, Zhao et al.~\cite{zhao2020tsasnet} provided a two-stage attention segmentation network, which can effectively alleviate the inhomogeneous intensity distribution problem by focusing on automatically catching the real tooth region. Lee et al.~\cite{lee2020application} adopted a fine-tuned mask R-CNN~\cite{he2017mask} algorithm to achieve the tooth segmentation. Nevertheless, these methods do not effectively solve the segmentation problem of fuzzy boundaries, and the performance improvement is mostly incremental. Chen et al.~\cite{chen2021mslpnet} proposed a novel MSLPNet with multi-scale structural similarity (MS-SSIM) loss, enhancing tooth segmentation that has fuzzy root boundaries. Cheng et al.~\cite{cheng2020learning} proposed U-Net+DFM to learn a direction field. It characterizes the directional relationship between pixels and implicitly restricts the shape of the segmentation result. Although these methods all achieve decent results in segmentation tasks, they are still limited by the intrinsic locality of convolutional neural networks (CNNs) and can not process global features very well. To relieve this problem, long-range dependencies via non-local operations are highly desired, and Transformer~\cite{vaswani2017attention} provides a modeling pipeline to achieve that. 
Chen et al. proposed TransUNet~\cite{chen2021transunet}, a Transformer-based encoder operating for segmentation, which adopts ViT~\cite{dosovitskiy2020image} with 12 Transformer layers as the encoder. However, ViT relies on pre-trained weights obtained by a huge image corpus, which results in undesirable performance on insufficient datasets. To solve that, Aravind et al.~\cite{srinivas2021bottleneck} presented BoTNet, an effective instance segmentation backbone, by combining transformer and convolution. Due to the high computational complexity of Transformer, BoTNet has only replaced a part of convolution in the last few layers of ResNet with Transformer. 

In order to mitigate the problems existing in the present approaches, our network GT U-Net employs both the combination of convolution and Transformer without pre-training weights and the grouping structure and bottleneck structure that significantly reduces the amount of computation. Besides, FD loss also solves the problem of fuzzy boundary segmentation by making full use of shape prior knowledge.
The main contributions of this paper are listed as follows:

(1) Our network retains the advantages of the general U-Net framework and introduces Transformer into the medical image segmentation application to solve the limitation of convolution.

(2) We design a grouping structure and a bottleneck structure, which greatly reduces the computation load of Transformer and makes it feasible in image segmentation.

(3) For the root segmentation task, we propose a shape-sensitive Fourier Descriptor loss function to deal with the problem of fuzzy boundary segmentation.

\begin{figure}[ht]
  \centering
  \includegraphics[width=1\textwidth]{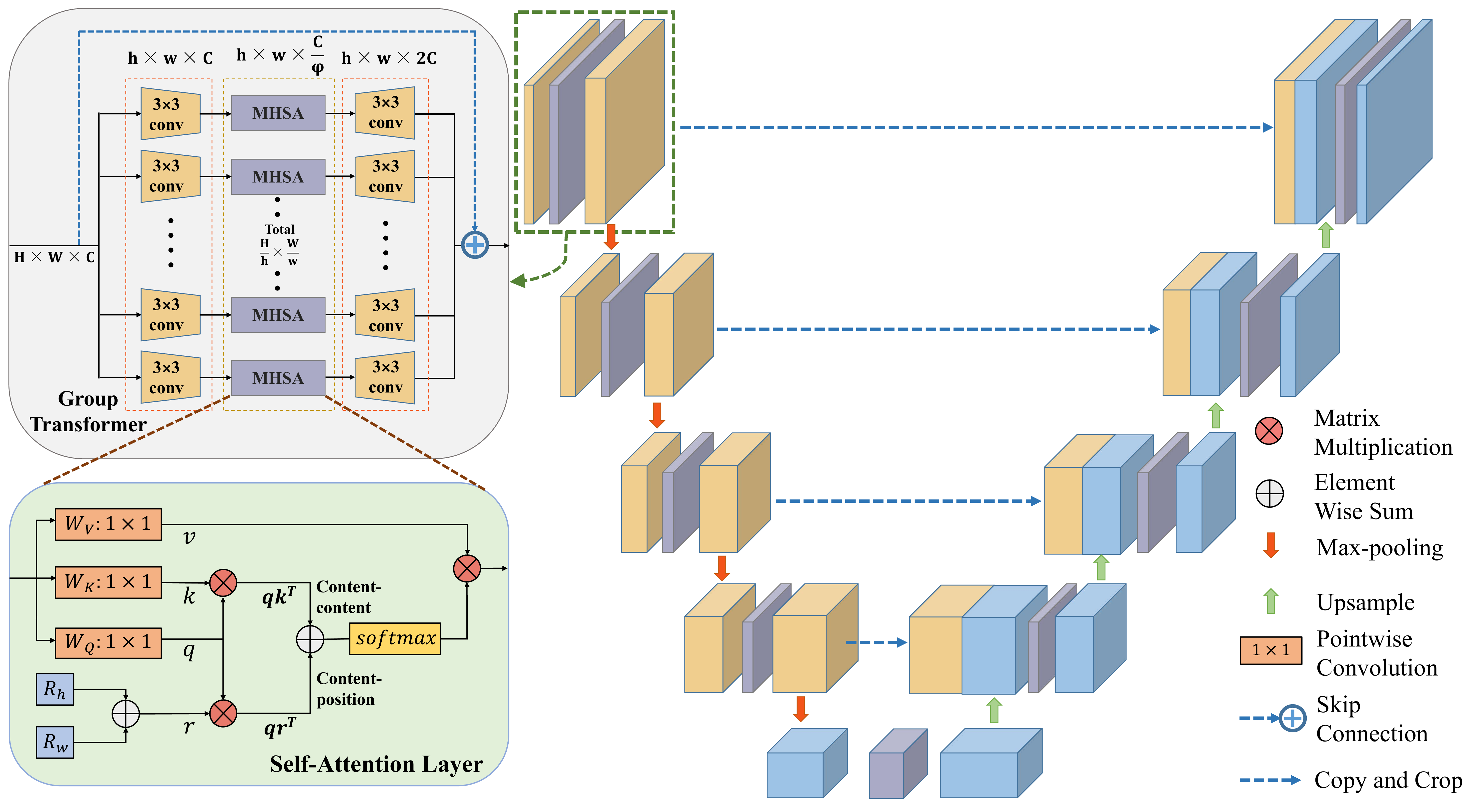}
  \caption{The architecture of the proposed GT U-Net, which is composed of a U-shape general framework and group Transformer. Self-attention layer is the base structure of Multi-head self-attention (MHSA) where our MHSA has 4 heads, and we do not illustrate them on the figure for simplicity.}
  \label{fig:fig2}
\end{figure}

\section{Method}
\label{sec:hints}
GT U-Net follows the overall U-shaped structure, where both the encoders and decoders consist of group Transformer. It works with a shape-sensitive Fourier Description (FD) loss function for tooth root segmentation. The proposed method is described in full detail below.

\subsection{U-Net like Group Transformer Network}
\textbf{Overall Architecture:} The architecture of GT U-Net is illustrated in Fig.~\ref{fig:fig2}. Specifically, our group Transformer is composed of a skip connection, a grouping module, $3\times 3$ convolutions, Multi-head self-attention (MHSA) modules, and a merging module. Among them, the skip connection is used to solve the problem of gradient vanishing and keep the low-level information. Considering that MHSA requires O($n^{2}d$) memory and computation when performing globally across $n$ entities, the grouping module and $3 \times 3$ convolutions are designed to reduce the amount of MHSA computation. \\\\
\textbf{Grouping Structure and Bottleneck Structure:} Transformer was first proposed for natural language processing (NLP) tasks and has become a hot topic in many computer vision tasks as a non-local operation with long-range dependence. However, it is difficult to apply Transformer directly to medical image tasks on account of the discrepancy between natural language and image. The number of words in natural language is limited, but the number of pixels increases quadratic with the increase of image size. Because of this, we designed group Transformer to solve the problem of too much computation due to the image characteristics in medical image segmentation. Assuming that the original feature block size is $H \times W \times C$, the computation of MHSA will be greatly reduced through the grouping structure and bottleneck structure. The calculation amount of MHSA before improvement~\cite{liu2021swin} and that of our group Transformer (GT) are given in Eq. (1) and (2), respectively:
$$
\Omega(MHSA) = 4HWC^2 + 2(HW)^2C  \eqno{(1)}
$$
$$
\Omega(GT) = 4hw(\frac{C}{\varphi})^2 + 2(hw)^2\frac{C}{\varphi} = \frac{4hw}{\varphi^2HW}HWC^2 + \frac{2h^2w^2}{\varphi H^2W^2}(HW)^2C  \eqno{(2)}
$$
where $\varphi$ is the channel scaling factor of the bottleneck structure, and the size of each group Transformer unit is set to $h \times w$, which is determined by the input image size and task characteristics. The proposed method achieves local long-range dependencies at higher encoder and decoder layers. With the deepening of the network, the field of vision is gradually expanded to extract the junction features between different GT units and to achieve the global long-range dependencies.\\\\
\textbf{Multi-Head Self-Attention:} MHSA~\cite{vaswani2017attention} is a type of attention mechanism which is more focused on the internal structure. The detailed architecture of MHSA is illustrated in Fig.~\ref{fig:fig2}, where the position encoding method is relative-distance-aware position encoding~\cite{shaw2018self,ramachandran2019stand,Bello_2019_ICCV} with $R_{h}$ and $R_{w}$ for height and width. The attention logit is $qk^T$ + $qr^T$, and $q, k, v, r$ denote query, key, value, and position encodings, respectively. \\\\
\textbf{Hybrid Structure of Convolution and Transformer:} 
Transformer has the ability to extract global contexts, but it also has the limitation of lacking translation invariance while locality and translational equivariance are the basic property of convolution. Therefore, $3\times 3$ convolution in our group Transformer is not only for the bottleneck structure, but it also helps construct a hybrid architecture where the convolution is responsible for feature extraction and Transformer is constructed to model the long-range dependencies.

\subsection{Shape-sensitive Fourier Descriptor Loss Function}
From the anatomical prior knowledge, it is known that the tooth roots share a similar shape. Adding the shape information into the loss function helps better guide the model to segment tooth roots. ($x_m,y_m$) is one of the coordinates on a tooth root boundary that contains $N$ pixels, and the boundary shape can be formed as a complex number $z(m)=x_m+jy_m$. The Fourier Descriptor~\cite{zahn1972fourier} of this shape is defined as $DFT(z(m))$, that is Z(k) in Eq. (3):
$$
Z(k)=DFT(z(m))=\frac{1}{N}\sum_{m=0}^{N-1}z(m)e^{-j2\pi mk/N}\;\;\;(k=0,\cdots,N-1)  \eqno{(3)}
$$
Fourier Descriptor is a quantitative representation of closed shapes independent of their starting point, scaling, location, and rotation. Therefore, the shape difference between the predicted boundary $A$ and the manually labeled boundary $B$ can be quantified, and it can be calculated based on Eq. (4). The original loss function is binary cross-entropy (BCE). In the proposed method, we add Fourier Descriptors to build a new shape-sensitive loss function. Considering their order of magnitude in the whole training process, we calculate the final Fourier Descriptor (FD) loss by Eq. (5):
$$
\Delta Z(k) =| Z_A(k) - Z_B(k)|.  \eqno{(4)}
$$
$$
FD\ loss = BCE(A,B) \times \frac{1}{1+e^{-\beta \times \Delta Z(k)}}  \eqno{(5)}
$$
The novel FD loss pays equal attention to shape loss and BCE loss regardless of their order of magnitude. $\beta$ is determined by the order of magnitude of $\Delta Z(k)$, and it is set to 10 in this paper.

\section{Experiment}
\subsection{Tooth Root Segmentation Dataset} 
We have built a new root canal therapy X-ray image dataset with patients' consent. The tooth root segmentation dataset contains 248 root canal therapy X-ray images from different patients in total. And three experienced stomatologists helped to complete the tooth root annotation. Specifically, to minimize inter-observer variability, the final annotation results needed the agreement of all three stomatologists.
\begin{figure}[ht]
  \centering
  \includegraphics[width=0.95\textwidth]{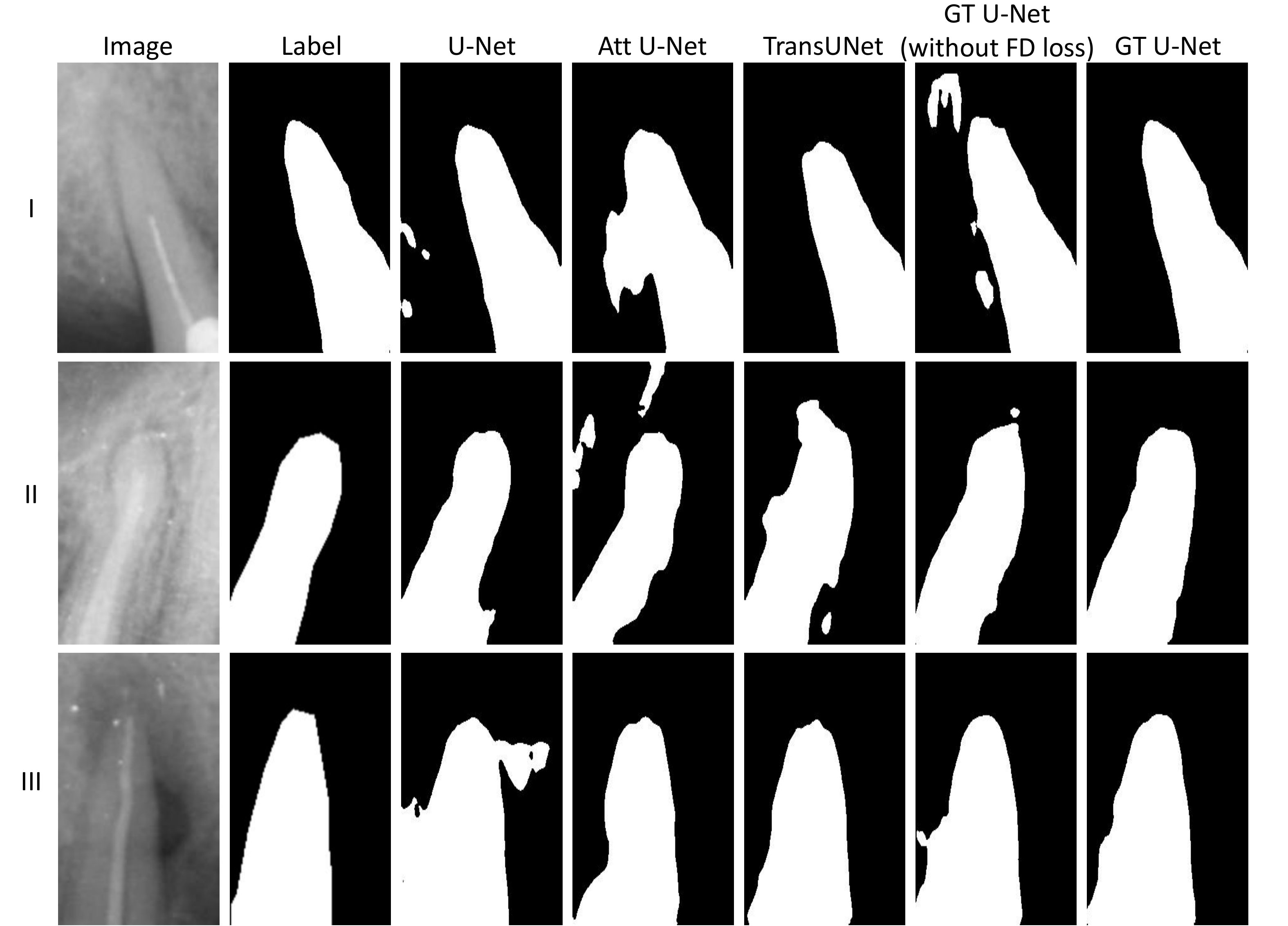}
  \caption{Representative segmentation results achieved by different methods.}
  \label{fig:toothres}
\end{figure}

\subsection{Implementation Details}
In our implementation, three state-of-the-art methods are compared, U-Net~\cite{ronneberger2015u}, Attention U-Net~\cite{oktay2018attention}, and TransUNet~\cite{chen2021transunet}. All compared methods and our method were implemented using PyTorch and trained on four RTX 2080Ti GPUs. Random center crop, random rotation, and axial flipping are adopted for the data augmentation to improve robustness and avoid overfitting. In the training, Adam optimizer is adopted and the initial learning rate and momentum are set to $2 \times 10^{-4}$ and 0.5, respectively. The total number of training epochs is 200, and the batch size is 12. All the images are resized to $256 \times 256$ for input. $h \times w$ of the group Transformer unit is set to $8 \times 8$, and $\varphi$ is set to 2.

To evaluate the segmentation performance, we compare the results with six metrics for evaluation, including Accuracy (ACC), Sensitivity (SE), Specificity (SP), Jaccard Similarity (JS), and Dice Coefficient (DICE). Besides, we adopt 3-fold cross-validation for testing the segmentation performance on the tooth root segmentation dataset.

\begin{table}[ht]
\caption{The evaluation performance of all methods on the Tooth Root Segmentation Dataset.}
  \label{tab:root}
  \centering
  \setlength{\tabcolsep}{1.4mm}
\begin{tabular}{c|c|c|c|c|c}
\hline
&U-Net &Att U-Net&TransUNet &\begin{tabular}[c]{@{}c@{}}\textbf{GT U-Net}\\(without FD loss)\end{tabular}&\textbf{GT U-Net}  \\ \hline
ACC(\%)& 92.81$\pm$0.90 &93.03$\pm$0.56&93.41$\pm$0.25&93.80$\pm$0.32 &\textbf{93.98$\pm$0.11} \\
SE(\%)& 92.37$\pm$0.68 &89.36$\pm$1.30&91.81$\pm$0.74&\textbf{93.98$\pm$0.38} &93.74$\pm$1.18 \\
SP(\%)& 93.74$\pm$0.99&\textbf{95.60$\pm$0.28}&95.05$\pm$0.43&94.26$\pm$0.35 &94.79$\pm$1.14 \\
JS(\%)&84.62$\pm$1.30&83.85$\pm$1.29&85.75$\pm$0.65&86.63$\pm$0.62 &\textbf{86.87$\pm$0.13} \\
DICE(\%) &91.19$\pm$1.01&91.15$\pm$0.76&91.96$\pm$0.40&92.32$\pm$0.49 &\textbf{92.54$\pm$0.25}  \\ \hline
\end{tabular}
\end{table}

\subsection{Experimental Results}
The average results with standard deviations of all methods are reported in Table~\ref{tab:root}. From this table, it can be summarized that our proposed GT U-Net achieves the best accuracy, sensitivity, Jaccard Similarity, and Dice Coefficient. In Fig.~\ref{fig:toothres}, we qualitatively compare the segmentation results on three oral X-ray images. It can be evidently seen that our method obtains the best segmentation results on the blurred boundaries. To verify the effectiveness of FD loss, we implement GT U-Net without FD loss. Based on the comparison in Table~\ref{tab:root} and Fig.~\ref{fig:toothres}, FD loss can be effectively applied to the segmentation task with a fuzzy boundary but a similar shape by implicitly restricting the shape of the segmentation result with the anatomical prior knowledge.

\begin{table}[ht]
\caption{The segmentation performance achieved by all methods on the public DRIVE dataset.}
  \label{tab:dirve}
  \centering
  \setlength{\tabcolsep}{2mm}
\begin{tabular}{c|c|c|c|c|c|c}
\hline
&Year&ACC(\%) &SE(\%) &SP(\%)  &F1(\%) & AUC(\%)    \\ \hline
U-Net~\cite{ronneberger2015u,jin2019dunet}&2015&95.55&78.22&98.08&81.74&97.52 \\ 
MS-NFN~\cite{wu2018multiscale} & 2018&95.67&78.44 &98.19 &- & 98.07 \\ 
Residual U-Net~\cite{alom2018recurrent,li2020iternet}&2018 &95.53&77.26&98.20&81.49&97.79 \\ 
Recurrent U-Net~\cite{alom2018recurrent,li2020iternet}&2018& 95.56&77.51&98.16&81.55&97.82 \\ 
R2U-Net~\cite{alom2018recurrent,li2020iternet}&2018& 95.56&77.92&98.13&81.71&97.84 \\
DenseBlock-UNet~\cite{li2018h}&2018& 95.41&79.28&97.76&81.46&97.56 \\
DUNet~\cite{jin2019dunet}&2019& 95.58&78.63&98.05&81.90&97.78 \\
ACE-Net~\cite{zhu2019ace}&2019&95.69&77.25&\textbf{98.42}&-&97.42 \\
IterNet~\cite{li2020iternet}&2020&95.73&77.35&98.38&82.05&\textbf{98.16} \\
\textbf{GT U-Net}&2021 &\textbf{96.31}&\textbf{82.54}&98.24&\textbf{84.58}&98.02 \\
\hline
\end{tabular}
\end{table}

\subsection{Performance on the Public DRIVE Dataset} 
To further evaluate the performance of our GT U-Net, we applied it on the widely used retinal dataset DRIVE~\cite{staal2004ridge}. The dataset consists of 40 randomly selected color fundus retinal images of size $565 \times 584$. Officially, DRIVE is split into two equal groups for training and testing. The group Transformer unit parameters $h$ and $w$ are all set to 4, and $\varphi$ is set to 2. Since FD loss is proposed for medical image segmentation with a similar shape, like tooth root segmentation, we did not implement it on the DRIVE dataset. To efficiently train our GT U-Net, we cropped image patches from the original images, and $64\times 64$ is adopted as the sample patch size, which is widely adopted on this dataset.

\begin{figure}[ht]
  \centering
  \includegraphics[width=0.9\textwidth]{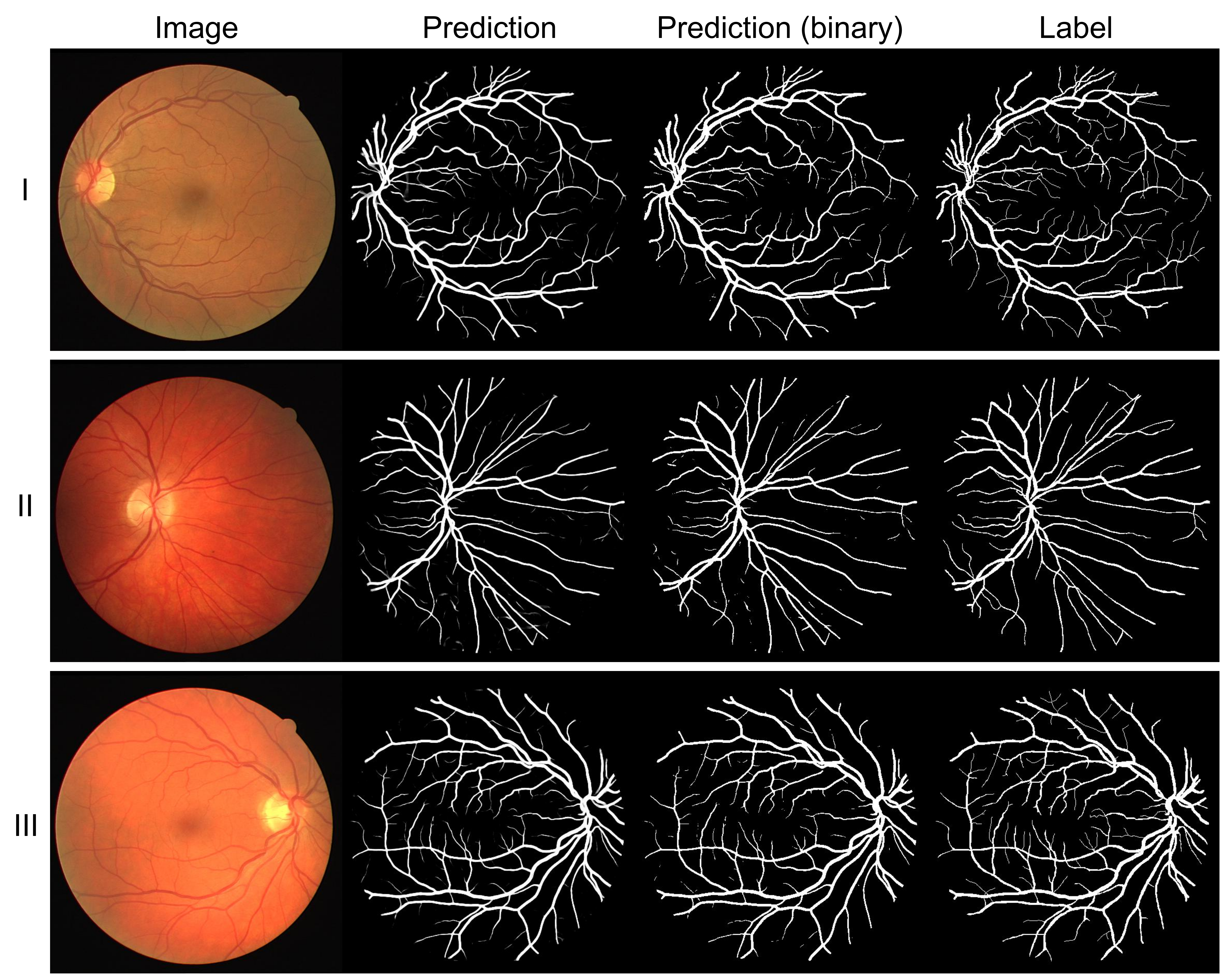}
  \caption{Representative segmentation results of the public DRIVE dataset.}
  \label{fig:driveres}
\end{figure}

The comparison results are reported in Table~\ref{tab:dirve}. It can be observed that our GT U-Net outperforms other methods by at least 3.2\% on SE, over 2.5\% on F1, and over 0.5\% on ACC. In terms of SP and AUC, our GT U-Net achieved competitive performance in comparison with the best methods, limiting the difference from them within 0.2\%. Meanwhile, the visualization of some representative segmentation results is given in Fig.~\ref{fig:driveres}. It can be clearly observed that the vascular part of the retina can be segmented very well by our method, except for some very tiny areas.

\section{Conclusion}
In this paper, a novel end-to-end U-Net like group Transformer network (GT U-Net) for medical image segmentation is presented. GT U-Net uses the hybrid structure of convolution and Transformer to eliminate the need for pre-trained weights. Moreover, it significantly reduces the computational complexity of Transformer using a bottleneck structure and a grouping structure, and FD loss can be effectively applied to the segmentation task with a fuzzy boundary but a similar shape. Extensive experiments are conducted on tooth root and retinal vessel segmentation tasks, respectively, and the results show that the proposed GT U-Net has excellent segmentation performance and a great application perspective in a wide range of medical image segmentation tasks.\\\\
\textbf{Acknowledgements.} This work was supported by the National Key Research and Development Program of China (Grant No. 2019YFC0118404) and Public Projects of Zhejiang Province (Grant No. LGG20F020001).

\bibliographystyle{unsrt}
\begingroup
  
  \small 
  \bibliography{paper}
\endgroup

\end{document}